\documentclass[conference]{IEEEtran}
\IEEEoverridecommandlockouts
\usepackage{cite}
\usepackage{amsmath,amssymb,amsfonts}
\usepackage{algorithmic}
\usepackage{graphicx}
\usepackage{textcomp}
\usepackage{xcolor}
\def\BibTeX{{\rm B\kern-.05em{\sc i\kern-.025em b}\kern-.08em
    T\kern-.1667em\lower.7ex\hbox{E}\kern-.125emX}}

\usepackage{booktabs}

\begin{document}

\title{Crowd Counting in Harsh Weather using Image Denoising with Pix2Pix GANs\\
\thanks{This publication was made possible by the PDRA award PDRA7-0606-21012 from the Qatar National Research Fund (a member of The Qatar Foundation) and Qatar University Internal Grant No. IRCC-2023-237. The statements made herein are solely the responsibility of the authors.}}

\author{Muhammad Asif Khan, \IEEEmembership{Senior Member, IEEE}, Ridha Hamila., \IEEEmembership{Senior Member, IEEE} and Hamid Menouar, \IEEEmembership{Senior Member, IEEE}}

\author{
  \IEEEauthorblockN{
  	Muhammad Asif Khan\IEEEauthorrefmark{1},
  	Hamid Menouar\IEEEauthorrefmark{1}, and
        Ridha Hamila\IEEEauthorrefmark{2},
  }

  \IEEEauthorblockA{
  Qatar Mobility Innovations Center, Qatar University. \IEEEauthorrefmark{1}
  	Department of Electrical Engineering, Qatar University\IEEEauthorrefmark{2}\\
        \{muhammada, hamidm\}@qmic.com\IEEEauthorrefmark{1}, 
        hamila@qu.edu.qa\IEEEauthorrefmark{2},
  }
  }

\maketitle

Visual crowd counting estimates the density of the crowd using deep learning models such as convolution neural networks (CNNs). The performance of the model heavily relies on the quality of the training data that constitutes crowd images. In harsh weather such as fog, dust, and low light conditions, the inference performance may severely degrade on the noisy and blur images. In this paper, we propose the use of Pix2Pix generative adversarial network (GAN) to first denoise the crowd images prior to passing them to the counting model. A Pix2Pix network is trained using synthetic noisy images generated from original crowd images and then the pretrained generator is then used in the inference engine to estimate the crowd density in unseen, noisy crowd images. The performance is tested on JHU-Crowd dataset to validate the significance of the proposed method particularly when high reliability and accuracy are required.

\begin{IEEEkeywords}
crowd counting, CNN, density estimation, GAN, Pix2Pix
\end{IEEEkeywords}

\section{Introduction}
Crowd density estimation is a task in computer vision that aims to estimate crowd density and counting using crowd images and videos. Crowd counting has applications in several domains such as public safety, event planning, transportation, security \& surveillance, resource allocation, urban planning, healthcare, and environmental monitoring. Although several methods have been applied traditionally \cite{Li_2008, Topkaya_2014, Chen_2012, Chan_2009}, convolution neural network (CNN) is the most widely used and effective approach. A CNN-based crowd-counting model is trained using a dataset of annotated crowd images. The images are point annotated by placing a dot on all head positions in the image. This produces a sparse localization map i.e., a matrix of zeros and ones where one represents the head position. To train the model, the localization map is transformed into a density map by applying a Gaussian kernel to the head position. The crowd model then regresses the target density maps. Over the years, several state-of-the-art crowd-counting CNN architectures are proposed \cite{khan2022revisiting}. Some examples include CrowdCNN \cite{CrowdCNN_CVPR2015}, MCNN \cite{MCNN_CVPR2016}, CMTL \cite{CMTL_AVSS2017}, CSRNet \cite{CSRNet_CVPR2018}, TEDnet \cite{TEDnet_CVPR2019}, SANet \cite{SANet_ECCV2018}, SASNet \cite{SASNet_AAAI2021}, MobileCount \cite{MobileCount_PRCV2019}, LCDnet \cite{Khan2023LCDnet}, and DroneNet \cite{Khan2022DroneNet}. These models are developed using numerous architectural improvements and training methods, resulting in incremental gains in counting accuracy over benchmark crowd counting datasets.
\par
Nevertheless, the prediction performance of these models degrades when the crowd images are blurry and noisy. Noisy crowd images are common in many practical scenarios such as during rain, fog, dust storms, object motion, etc. When high accuracy is desired, it is inevitable that the crowd model is highly robust against such image distortions. To provide highly accurate counting performance, this work proposes to enhance the crowd image prior to passing it through the counting model. Although traditional image enhancement methods can be used for image denoising, more powerful methods such as generative adversarial networks (GANs) \cite{GANs_2014} offer superior capabilities for image enhancement over several image attributes (such as brightness, pixel distortion, illumination, contrast, etc.) and produces a clearer and enhanced version of the noisy crowd image.
Pix2Pix \cite{Pix2Pix_2017} is a general-purpose image translation GAN model used in many tasks such as map-to-aerial photos, black and white-to-color photos, sketch-to-color photos, day-to-night photos, thermal-color, semantic-to-photos, etc. This paper proposes the use of Pix2Pix GANs for crowd image denoising and enhancement prior to passing to the crowd model for density estimation. Pix2Pix GANs have been applied previously in various image translation tasks \cite{deLima2022Pix2PixNT, Toda2022LungCC, Aljohani2022} with significant improvements in results.

The contribution of this paper is as follows:
\begin{itemize}
\item A two-stage crowd density estimation method is proposed for applications requiring high accuracy and robustness against extreme weather and environmental conditions using integrating a Pix2Pix GAN and a density estimation model.
\item A general-purpose Pix2Pix GAN network is trained over crowd datasets of pairs of (clear and noisy) images. We synthetically added several noise and distortion effects to mimic real-world harsh weather scenarios.
\item The pretrained generator of the Pix2Pix GAN model is then used as a preprocessor in several crowd density estimation models. The prediction performance of the GAN-aided crowd-counting models is compared with the standard counting crowd models to validate the significance of the proposed scheme.
\end{itemize}

\section{Related Work}

Crowd counting using convolution neural networks (CNNs) was first introduced in \cite{CrowdCNN_CVPR2015} using a simple architecture. Following this, several lightweight architectures were proposed in \cite{MCNN_CVPR2016, MSCNN_ICIP2017, CMTL_AVSS2017, CrowdNet_CVPR2016}. These models achieved incremental gains in accuracy over benchmark datasets. However, significant gains were achieved using crowd architectures using pretrained frontend networks. These include CSRNet \cite{CSRNet_CVPR2018}, CANNet \cite{CANNet_CVPR2019}, GSP \cite{GSP_CVPR2019}, Deepcount \cite{DeepCount_ECAI2020}, SASNet, and SGANet \cite{SGANet_IEEEITS2022}.

Generative adversarial network (GAN) was first introduced in \cite{GANs_2014} as a two-stage network that is comprised of two neural networks, a generator network, and a discriminator network. The generator is trained to produce images in the target domain, and the discriminator judges these images and classifies them as real or fake. The end goal of the GAN is to make the generator output as realistic (similar to the target images) as possible. There have been several types of GAN architectures such as conditional GANs (cGAN) \cite{cGAN_2014}, DCGANs \cite{DCGAN_2016}, CycleGANs \cite{CycleGAN_2017}, StyleGAN \cite{StyleGAN_2021}, and Pix2Pix GAN \cite{Pix2Pix_2017}.
\par
Pix2Pix GANs are general-purpose GANs used in various image translation tasks such as sketch-to-photograph conversion \cite{Raghavendra2022}, image dehazing \cite{Qu2019EnhancedPD}, depth image generation \cite{Shimada2022}, and computerized tomography (CT) scan image generation \cite{Toda2022LungCC}. They have been used in image denoising in several studies. Authors in \cite{raposoultrasound} use a Pix2Pix GAN to denoise ultrasound images. Speckle noise using Rayleigh distribution was added to the training images to generate image pairs of original and noisy images.
Authors in \cite{Li2021AnIP} use a Pix2Pix GAN with a penalty term to render greyscale images into color images. The performance is evaluated over PSNR and SSIM metrics. A pretrained Pix2Pix GAN is used in \cite{Jadhav2022Pix2PixGA} to denoise degraded electronic documents. The architecture replaces the U-net network in the generator with ResNet6, and patchGAN in the discriminator network.
Similarly, \cite{Jiang2022AGI} uses a Pix2Pix GAN with a residual network-based generator and a PatchGAN-based discriminator to enhance CT scan images. Other works on using Pix2Pix GAN for image denoising include \cite{Park2019UnpairedID, Kalpana2022Pix2PixGI, Sun2021Pix2PixGA}.

\section{Proposed Scheme}
We used a Pix2Pix GAN which denoises blurry crowd images to acquire clear images which are then passed to a crowd density estimation model. The Pix2Pix GAN is trained using pair of noisy (input) and clear (target) crowd images

\subsection{The Pix2Pix GAN architecture}
The Pix2Pix GAN consists of a generator and a discriminator network. The generator consists of a UNet architecture. Both the generator and the discriminator consist of Convolution-Batch Normalization-ReLU blocks.
\begin{figure}[!h]
    \centering
    \includegraphics[width=0.9\columnwidth]{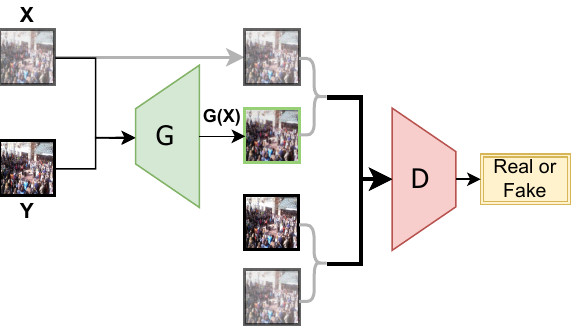}
    \caption{Image Denoising using Pix2Pix GAN. The images with gray borders are input images, images with black borders are target images, and the image with the green border is the generator's prediction. \textbf{G} is the generator and \textbf{D} is the discriminator of the Pix2Pix GAN.}
    \label{fig:pix2pix}
\end{figure}
The generator network translates the noisy image into an enhanced image as the input image. The discriminator attempts to classify the generator output as real or fake. The generator network uses$L_1$ or $L_2$ loss function as follows:

\begin{equation}
\mathcal{L}_{\text{L1}}(G) = \frac{1}{N} \sum_{i=1}^{N} \left| G(x_i) - y_i \right|
\label{eq:l1_loss}
\end{equation}
where $G$ is the generator network, $N$ is number of samples, $x_i$ is the input sample, and $y_i$ is the ground truth value. 

The discriminator network uses binary cross entropy (BCE) loss or adversarial loss to measure the similarity between the discriminator's predictions for the generator's output and the target image. It is calculated as Eq. \ref{eq:bce_loss}:

\begin{equation}
\begin{aligned}
\mathcal{L}_{\text{BCE}}(D, G) &= - \frac{1}{N} \sum_{i=1}^{N} \bigg( y_i \log D(x_i) \\
&\quad + (1 - y_i) \log (1 - D(G(x_i))) \bigg)
\end{aligned}
\label{eq:bce_loss}
\end{equation}

where $D$ is the discriminator network, $N$ is the total number of samples, $D_{(x_i)}$ is discriminator's prediction for input $x_i$, and $D(G_{(x_i)})$ is discriminator's prediction for the generated output $G_{(x_i)}$.

\section{Experiments and Results}

\subsection{Datasets and Baselines}
\begin{figure}[!h]
    \centering
    \includegraphics[width=0.8\columnwidth]{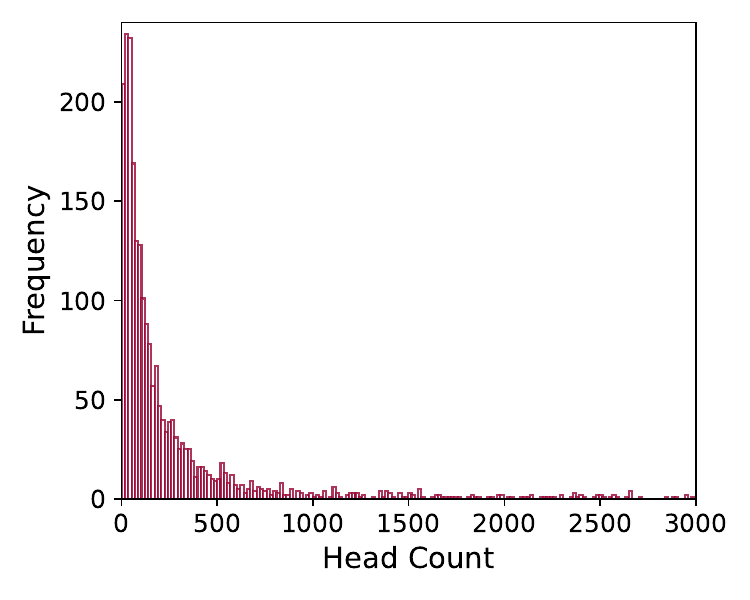}
    \caption{Histogram of head counts in the JHU-Crowd dataset.}
    \label{fig:jhu_hist}
\end{figure}

\begin{figure}[!h]
    \centering
    \includegraphics[width=0.95\columnwidth]{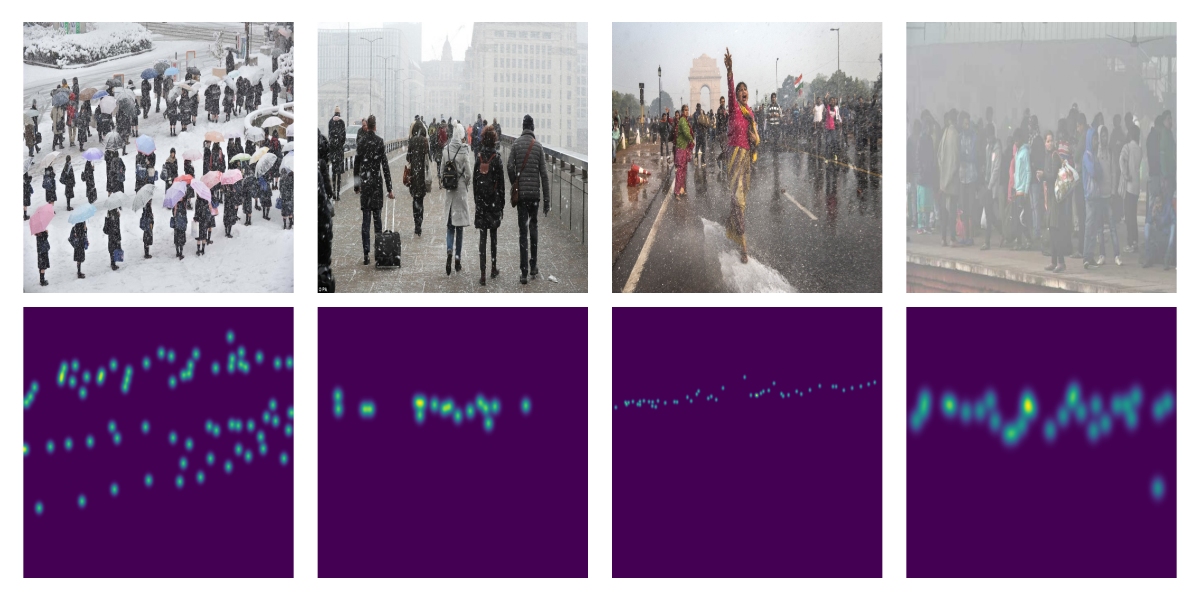}
    \caption{Noisy image samples from JHU-Crowd dataset showing harsh weather conditions.}
    \label{fig:jhu_samples}
\end{figure}

To train the Pix2Pix GAN, we use the ShanghaiTech dataset \cite{MCNN_CVPR2016}. The dataset contains a total of 1198 images and 330,165 annotations. The dataset is divided into two parts. Part A contains 482 images and part B contains 716 images.
Fig. \ref{fig:jhu_hist} presents the histogram of per-image counts in the JHU-Crowd++ whereas Fig. \ref{fig:jhu_samples} shows some representative samples from the subset.

As pairs of images (input and target images) are required to train the GAN, random image distortions are applied to the original images such that the original image is the target image whereas the distorted image is the input image. The pretrained model is then applied to JHU-Crowd++ \cite{jhucrowd_dataset2020} dataset to estimate crowd counts. The JHU-Crowd++ dataset contains 4,372 images and 1.51 million annotations. The dataset further contains additional attributes such as adverse weather and illumination conditions. We used only the images with adverse weather conditions to test the proposed method in this paper. 
\par
The performance is validated using four crowd counting models including MCNN \cite{MCNN_CVPR2016}, CMTL \cite{CMTL_AVSS2017}, CSRNet \cite{CSRNet_CVPR2018}, TEDnet \cite{TEDnet_CVPR2019}, and SANet \cite{SANet_ECCV2018}. MCNN is a lightweight multicolumn network consisting of three columns with a fusion layer at the end. CSRNet is a deep architecture using the pretrained VGG-16 frontend (first 10 layers). TEDnet is an encoder-decoder network. SASNet is a scale-aggregation network using a composite loss function i.e., a combination of Euclidean loss and local pattern consistency loss.

\subsection{Evaluation Metrics}
Structural similarity index measure (SSIM), and pixel signal-to-noise ratio (PSNR) are two standard metrics to compare the performance of image translation using GANs (including Pix2Pix) model. PSNR measures the difference between the corresponding pixels in the target image (label) and the predicted image using GAN. SSIM measures the difference between corresponding pixels in terms of brightness, contrast, and structure. These are defined as follows:
\begin{equation} \label{eq:ssim}
    SSIM (x,y) = \frac{(2\mu_x \mu_y + C_1)  (2\sigma_x \sigma_y C_2)}  {(\mu_z^2 \mu_y^2 + C_1)  (\mu_z^2 \mu_y^2 + C_2)}
\end{equation}
where $\mu_x, \mu_y, \sigma_x, \sigma_y$ represent the means and standard deviations of the actual and predicted density maps, respectively.
\begin{equation} \label{eq:psnr}
    PSNR = 10 log_{10}\left( \frac{Max(I^2)}{MSE}  \right)
\end{equation}
where $Max(I^2)$ is the maximal in the image data. If it is an 8-bit unsigned integer data type, the $Max(I^2)=255$.

To measure the counting performance, we use the standard metric i.e., the mean absolute error (MAE) which is calculated as follows:
\begin{equation} \label{eq:mae}
    MAE = \frac{1}{N} \sum_{1}^{N}{(e_n - \hat{g_n})}
\end{equation}

\subsection{Training}
The Pix2Pix GAN model is trained for 500 epochs using a batch size of 32 and the Adam optimizer with learning rates of $2e-4$ for both generator and discriminator networks. A dropout of 0.5 was used in some layers of the generator network. The crowd models are trained for 200 epochs using a batch size of 16 and the Adam optimizer with a base learning rate of $0.0001$.

\subsection{Results and Discussion}
This section presents the experimental results of Pix2Pix GAN for image denoising and the crowd counting performance using the baseline models and our proposed scheme. The Pix2Pix GAN model trained on ShanghaiTech dataset is evaluated on the JHU-Crowd++ dataset.

\begin{figure*}[!h]
    \centering
    \includegraphics[width=0.8\textwidth]{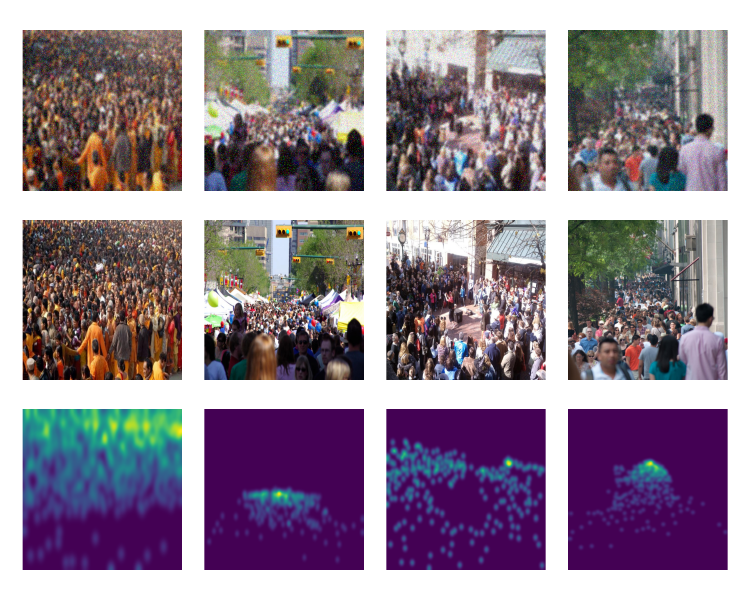}
    \caption{Samples from training data (ShanghaiTech Part A) dataset for Pix2Pix GAN. The top row contains noisy images, the middle row contains target images, and the bottom row contains the corresponding density maps (density maps are not used in GAN training).}
    \label{fig:gan_train_data}
\end{figure*}

Fig. \ref{fig:gan_train_data} shows samples from the training data for the Pix2Pix GAN. The training images are synthetically generated by applying various noise types (Gaussian noise, Gaussian blurs, etc.) to original images (serve as labels for the GAN).

\begin{figure}
    \centering
    \includegraphics[width=0.9\columnwidth]{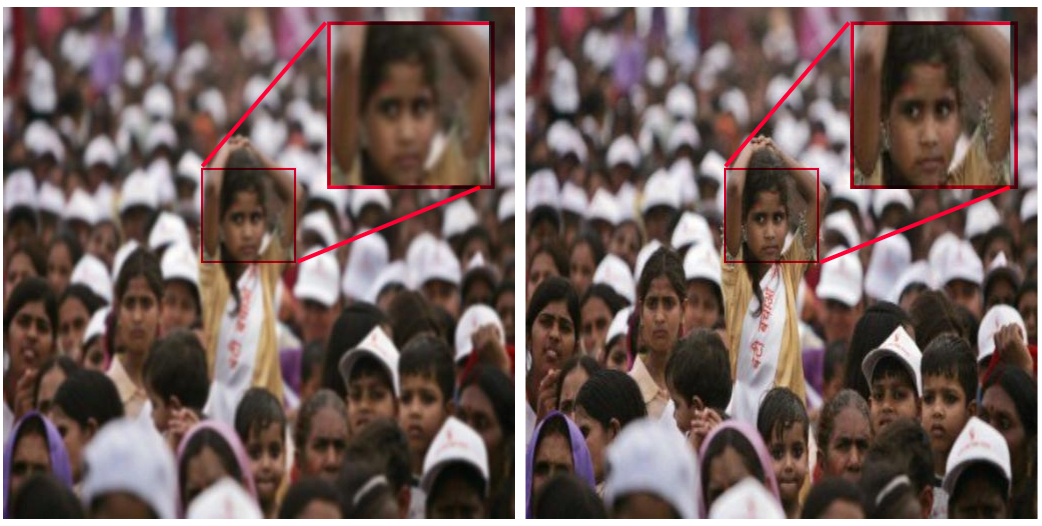}
    \caption{Illustration of image denoising using Pix2Pix GAN. The image on the left shows the input image (noisy) whereas the image on the right is the output image (GAN output).}
    \label{fig:gan_result}
\end{figure}

Fig. \ref{fig:gan_result} shows a visual illustration of image denoising on a single example image before and after passing through the GAN model. The difference in the noise level is very clear when zoomed in.

\begin{figure}[!h]
    \centering
    \includegraphics[width=0.45\columnwidth]{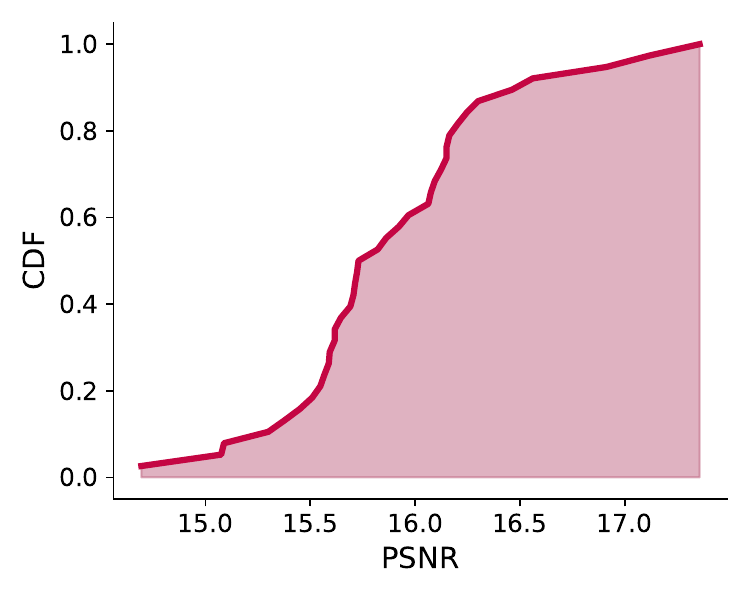}
    \includegraphics[width=0.45\columnwidth]{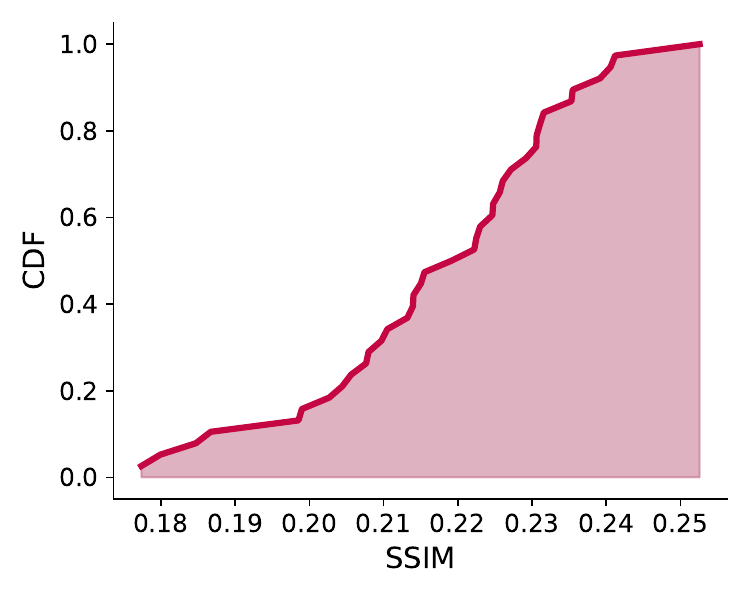} \\
    \includegraphics[width=0.45\columnwidth]{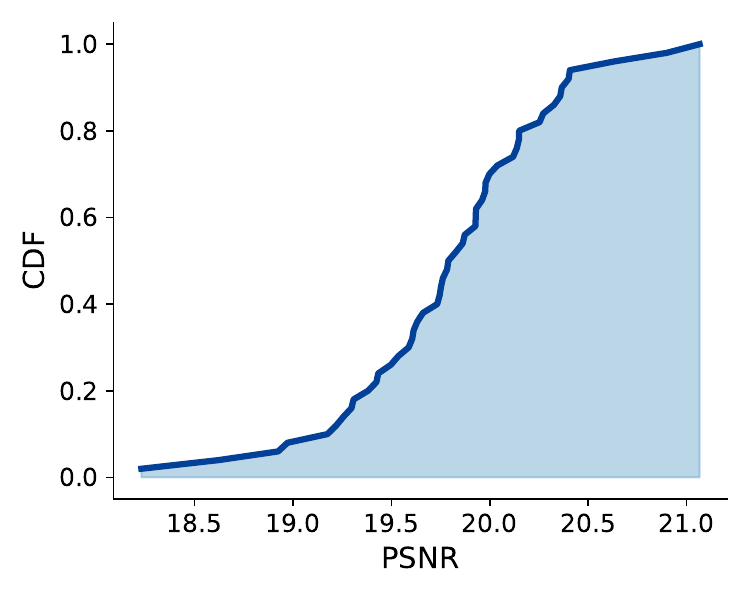}
    \includegraphics[width=0.45\columnwidth]{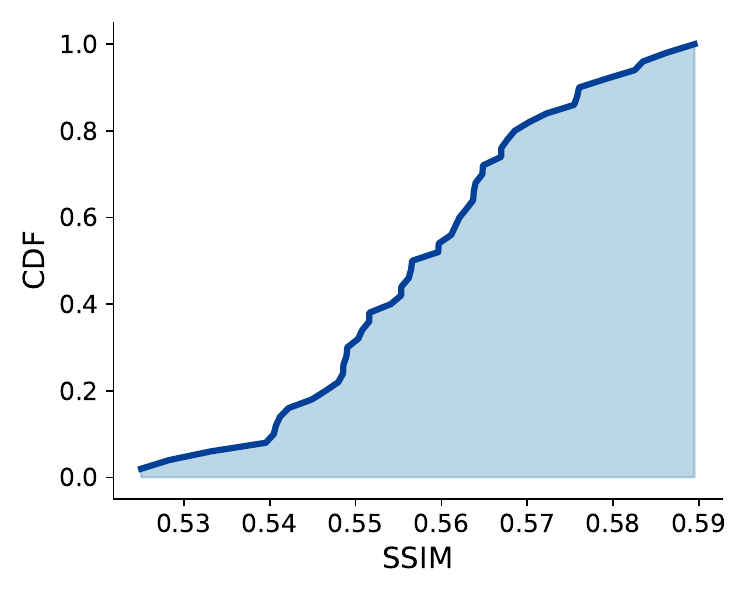}
    \caption{CDF of PSNR (left) and SSIM (right) values of the images in the training set (top) and the generated images using Pix2Pix GAN (bottom).}
    \label{fig:ssim_psnr}
\end{figure}

The performance over the entire dataset is then evaluated using the PSNR and SSIM metrics. The CDF of the two metrics is shown in Fig. \ref{fig:ssim_psnr}. The top two subplots show the CDF curves before applying the GAN model, whereas the bottom two CDF subplots show the results after applying the GAN model to the images. The difference is very clear in image quality enhancements as depicted in the figure e.g., the PSNR values first ranged between 14.5 and 18.0, which after applying the GAN ranged between 18.5 and 21.0. Similarly, the SSIM range was 0.18-0.25 (before applying GAN) which is improved to 0.53-0.59 (after applying the GAN model), indicating major improvements in the image quality.
\par

\begin{figure}
    \centering
    \includegraphics[width=0.9\columnwidth]{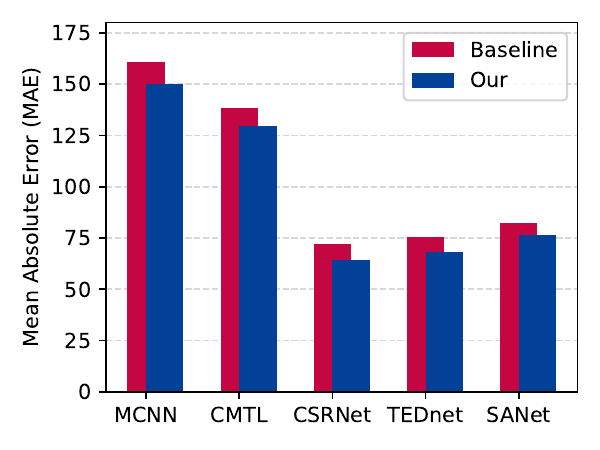}
    \caption{Performance comparison of various crowd counting models using Mean Absolute Error (MAE) metric.}
    \label{fig:counting}
\end{figure}

The trained Pix2Pix model is then applied to the JHU-Crowd++ dataset to enhance noisy images in the dataset. The two versions of the JHU-Crowd++ dataset i.e., original (with noisy images) and enhanced (without noisy images) are then used to train crowd-counting models. The baselines for our analysis are the crowd models trained on the original dataset. The baseline models are trained until the loss converges (did not reduce further after 10 epochs). Then the pretrained models are further trained (finetuned) on the enhanced dataset (containing images generated with the GAN model). The counting performance is evaluated using the mean absolute error (MAE) metric in Eq. \ref{eq:mae} for both methods i.e., models trained on the original dataset (Baselines), and models trained on the enhanced dataset (ours).
The performance of our proposed method is compared against all baselines )i.e., MCNN, CMTL, CSRNet, TEDnet, and SANet) and the results are reported in Fig. \ref{fig:counting}. The figure shows that the MAE values for all models using our proposed method are lower (lower MAE means better counting accuracy) than the baselines (without image denoising).

\section{Conclusion}
In this paper, we propose a two-stage crowd density estimation and counting model for harsh weather conditions using generative adversarial networks. A Pix2Pix GAN is applied to create enhanced and clear versions of the blurry and noisy crowd images taken in harsh weather such as fog, rain, and dust causing distortions. The enhanced images are then passed to the crowd model for density estimation. Experiments conducted on the JHU-Crowd dataset using several mainstream crowd models show superior performance as compared to the baseline over several metrics. It is worth noting that during the inference, the proposed method incurs extra delay and thus it may not be suitable for real-time performances. However, it can be a good choice for applications requiring high reliability and accuracy.

\bibliographystyle{ieeetr}
\bibliography{biblio}

\end{document}